\documentclass[10pt,twocolumn,letterpaper]{article}

\usepackage[pagenumbers]{iccv} 

\usepackage{amsmath,amsfonts}
\usepackage{amssymb}
\usepackage{bbm}
\usepackage{multirow}
\usepackage[linesnumbered,ruled,vlined]{algorithm2e}
\usepackage{colortbl}
\usepackage{xcolor} 

\definecolor{iccvblue}{rgb}{0.21,0.49,0.74}
\usepackage[pagebackref,breaklinks,colorlinks,allcolors=iccvblue]{hyperref}

\def\paperID{3102} 
\def\confName{ICCV}
\def\confYear{2025}

\title{Partial Forward Blocking: A Novel Data Pruning Paradigm for Lossless Training Acceleration}

\author{
    Dongyue Wu, Zilin Guo, Jialong Zuo, Nong Sang, Changxin Gao\thanks{Corresponding author.}\\
    National Key Laboratory of Multispectral Information Intelligent Processing Technology, \\School of Artifcial Intelligence and Automation, Huazhong University of Science and Technology\\
    \{dongyue\_wu, zilin\_guo, jlongzuo, nsang, cgao\}@hust.edu.cn
}

\begin{document}
\maketitle
\begin{abstract}
The ever-growing size of training datasets enhances the generalization capability of modern machine learning models but also incurs exorbitant computational costs. Existing data pruning approaches aim to accelerate training by removing those less important samples. However, they often rely on gradients or proxy models, leading to prohibitive additional costs of gradient back-propagation and proxy model training.
In this paper, we propose Partial Forward Blocking (PFB), a novel framework for lossless training acceleration. The efficiency of PFB stems from its unique adaptive pruning pipeline: sample importance is assessed based on features extracted from the shallow layers of the target model. 
Less important samples are then pruned, allowing only the retained ones to proceed with the subsequent forward pass and loss back-propagation. This mechanism significantly reduces the computational overhead of deep-layer forward passes and back-propagation for pruned samples, while also eliminating the need for auxiliary backward computations and proxy model training.
Moreover, PFB introduces probability density as an indicator of sample importance. 
Combined with an adaptive distribution estimation module, our method dynamically prioritizes relatively rare samples, aligning with the constantly evolving training state.
Extensive experiments demonstrate the significant superiority of PFB in performance and speed.
On ImageNet, PFB achieves a 0.5\% accuracy improvement and 33\% training time reduction with 40\% data pruned. 

\end{abstract}    
\section{Introduction}
\label{sec:intro}

The availability of large-scale datasets\cite{russakovsky2015imagenet,kirillov2023segment,changpinyo2021conceptual, abu2016youtube, netzer2011reading} in recent years has propelled significant advancements in deep learning.
Many powerful vision models\cite{radford2021learning,rombach2022high,kirillov2023segment} rely on vast amounts of training data to achieve robust generalization.
However, behind the unprecedented generalization performance lies a critical challenge: the enormous volume of training samples substantially increases the computational cost of training.
Thus, reducing the training cost has become a key and pressing concern.

\begin{figure}[t]
  \centering
   \includegraphics[width=\linewidth]{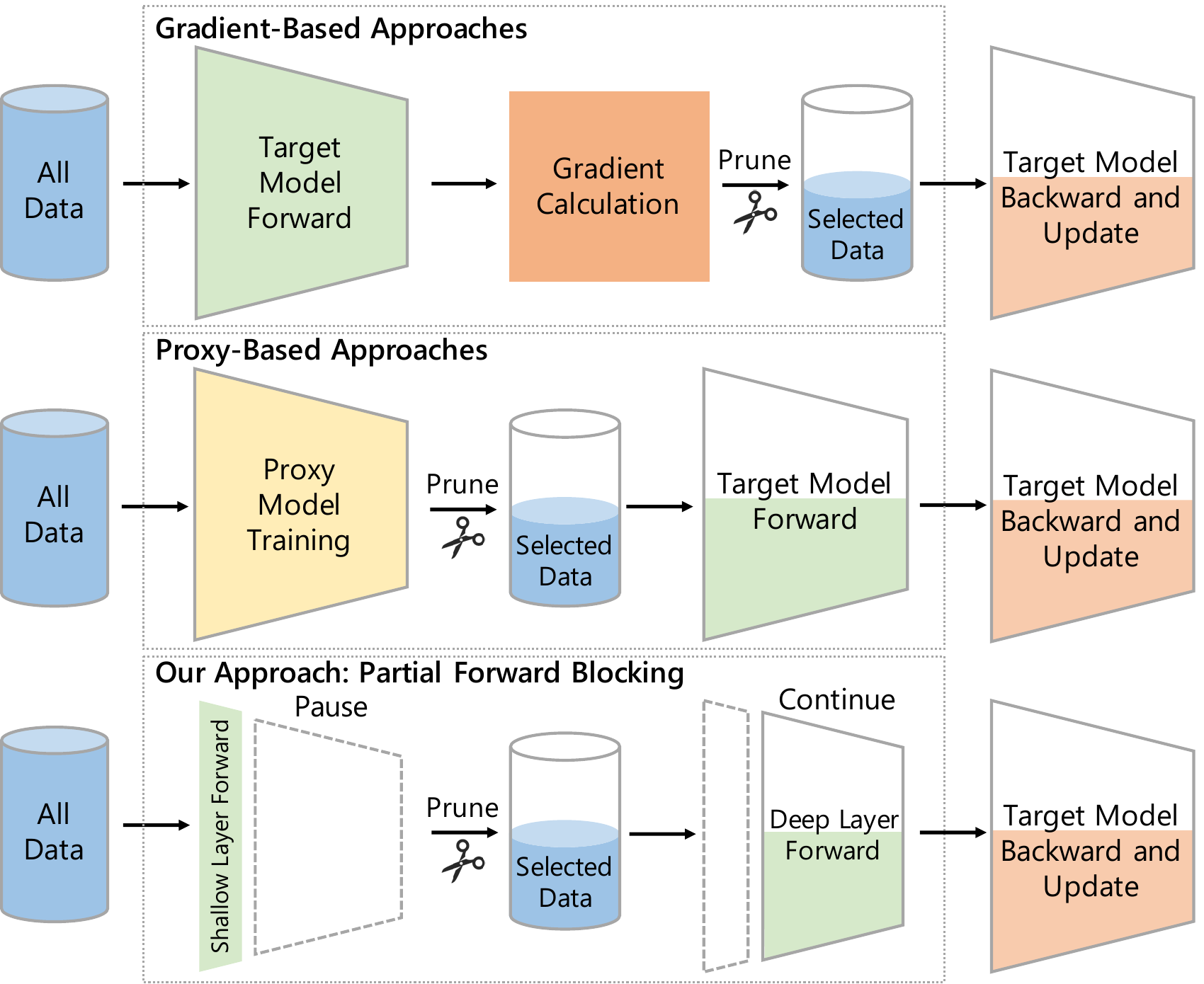}

   \caption{Computation cost comparison between Partial Forward Blocking~(PFB) and existing approaches. 
   The half blanks in \textbf{Forward} and \textbf{Update} indicate that only the selected samples participate in the network's forward inference, back-propagation, and updates. The proposed PFB strategy avoids the additional cost of gradient calculation and proxy model training by pruning based on features of the shallow network and blocking the subsequent deep network forward pass of the pruned samples.}
   \vspace{-1.3em}
   \label{fig:fig1}
\end{figure}

Replacing large datasets with a compact subset offers an intuitive and effective solution.
Dataset distillation\cite{zhao2023dataset,cazenavette2022dataset,wang2022cafe,nguyen2020dataset, zhaodataset} aims to synthesize a small yet highly informative set of samples by leveraging information from the original dataset.
However, the synthetic data often deviates significantly from real-world samples.
In contrast, data pruning\cite{svp,sorscher2022beyond,EL2N,dataset_pruning} and online batch selection\cite{loshchilov2015online,rho,DivBS} directly selects a subset of real samples, ensuring authenticity.
Yet, these existing methods typically rely on additional proxy models\cite{svp,deng2024,rho} or computationally expensive metrics, such as gradients\cite{EL2N,dataset_pruning,moso,DivBS}, to assess sample importance.
The computational burden introduced by proxy models, along with the cost of computing losses and back-propagating gradients, becomes increasingly prohibitive as dataset size grows, thereby undermining their efficiency.

This raises a critical question: \textit{Can we efficiently select training samples at a lower cost while preserving the strong generalization of target models?}
We believe an effective solution to this problem should possess the following characteristics:  1) \textbf{Efficiency}: Its computational overhead of importance evaluation should be minimal, making it feasible for large-scale datasets.
2) \textbf{Diversity}: The selected samples should span a broad and informative distribution to ensure a strong generalization and powerful performance.
3) \textbf{Adaptability}: The method should dynamically adjust to the evolving training state of the model.

This paper proposes a novel paradigm that significantly reduces training costs while preserving model performance to achieve these goals. 
We introduce the Partial Forward Blocking (PFB) strategy, which prunes samples in a batch-wise manner at the early stage of the forward pass to eliminate unnecessary computation as shown in \cref{fig:fig1}.
First, the target network is partitioned into a shallow sub-network and a deeper subsequent one.
Importance scores are computed based on the features from the shallow sub-network.
Samples with low importance are discarded immediately, blocking their forward pass through the deeper layers, thereby reducing computation costs.
Meanwhile, the retained samples reuse their extracted features to continue forward propagation, further improving efficiency.
Compared with gradient-based and proxy-based works, the PFB strategy avoids the substantial overhead of gradient back-propagation and proxy model training for importance scoring and significantly reduces the forward cost of the pruned samples.

To ensure the diversity of retained samples, we introduce a new importance metric, which leverages probability density as an indicator of sample redundancy: samples with high probability density are common and less informative, making them less important, while those with low probability density are rare and valuable for generalization. 
This criterion ensures the diversity of the retained subset by prioritizing rare samples and preventing excessive concentration in densely populated feature regions.
Additionally, an Adaptive Distribution Estimation~(ADE) module continuously updates its estimated feature distribution based on retained samples, enabling PFB to dynamically adjust to the evolving training state. 
By adaptively favoring rare and informative samples, PFB ensures that the retained subset remains well-aligned with the model’s learning needs.
Extensive experiments demonstrate that PFB achieves superior training efficiency with minimal computational overhead while maintaining competitive performance.
For example, PFB prunes 40\% training samples of ImageNet-1k but still achieve a 0.5\% accuracy improvement and 33.2\% training time reduction over the full data baseline.
In summary, our contributions are as follows:
\begin{enumerate}
    \item We propose an efficient training acceleration pipeline that conducts pruning in a batch-wise manner at the early stage of forward and blocks the subsequent forward pass of pruned samples.
    \item To ensure the diversity of selected samples, we propose utilizing the probability density of the feature distribution to measure importance.
    \item We introduce a distribution estimation module that adaptively tracks the distribution shift during training.
\end{enumerate}

\section{Related Work}
\label{sec:related}

Extracting a small amount of valuable samples from large datasets is a common objective in many other tasks, such as active learning\cite{bordes2005fast,sener2017active}, noisy learning\cite{mirzasoleiman2020coresets}, curriculum learning\cite{yoon2022online}, and data distillation\cite{wang2022cafe,nguyen2020dataset}. Since our approach focuses on accelerating training using real data, we primarily introduce data pruning and online batch selection.

\subsection{Data Pruning}
Data Pruning aims at selecting a small subset of training samples that can achieve comparable results as the original dataset to accelerate model training. 
These methods can be divided into two types: static and dynamic data pruning. 

Static methods only prune the dataset once. Thus, the performance of target models trained on the pruned datasets demonstrates the effectiveness of static methods.
The pioneering work \textit{Herding}~\cite{herding} tends to keep samples that are close to class centers. 
EL2N~\cite{EL2N} quantifies the sample-wise learning difficulty by averaging the norm of error of a set of networks.
Similarly, GraNd~\cite{EL2N} adopts the expectation of the gradient norm as an importance indicator.
\textit{Dataset Pruning}~\cite{dataset_pruning} and MoSo~\cite{moso} prune the samples which do the least damage to the empirical risk based on back-propagated gradients.
SVP~\cite{svp} and YOCO~\cite{yoco} both select samples according to the entropy and error of the prediction by pre-trained proxy models, respectively.
On the other hand, \textit{Forgetting}~\cite{fogetting} measures the learning difficulty via the forgetting times during training, namely the change of prediction from correct to incorrect.

\begin{figure*}[ht!]
  \centering
  \includegraphics[width=0.9\linewidth]{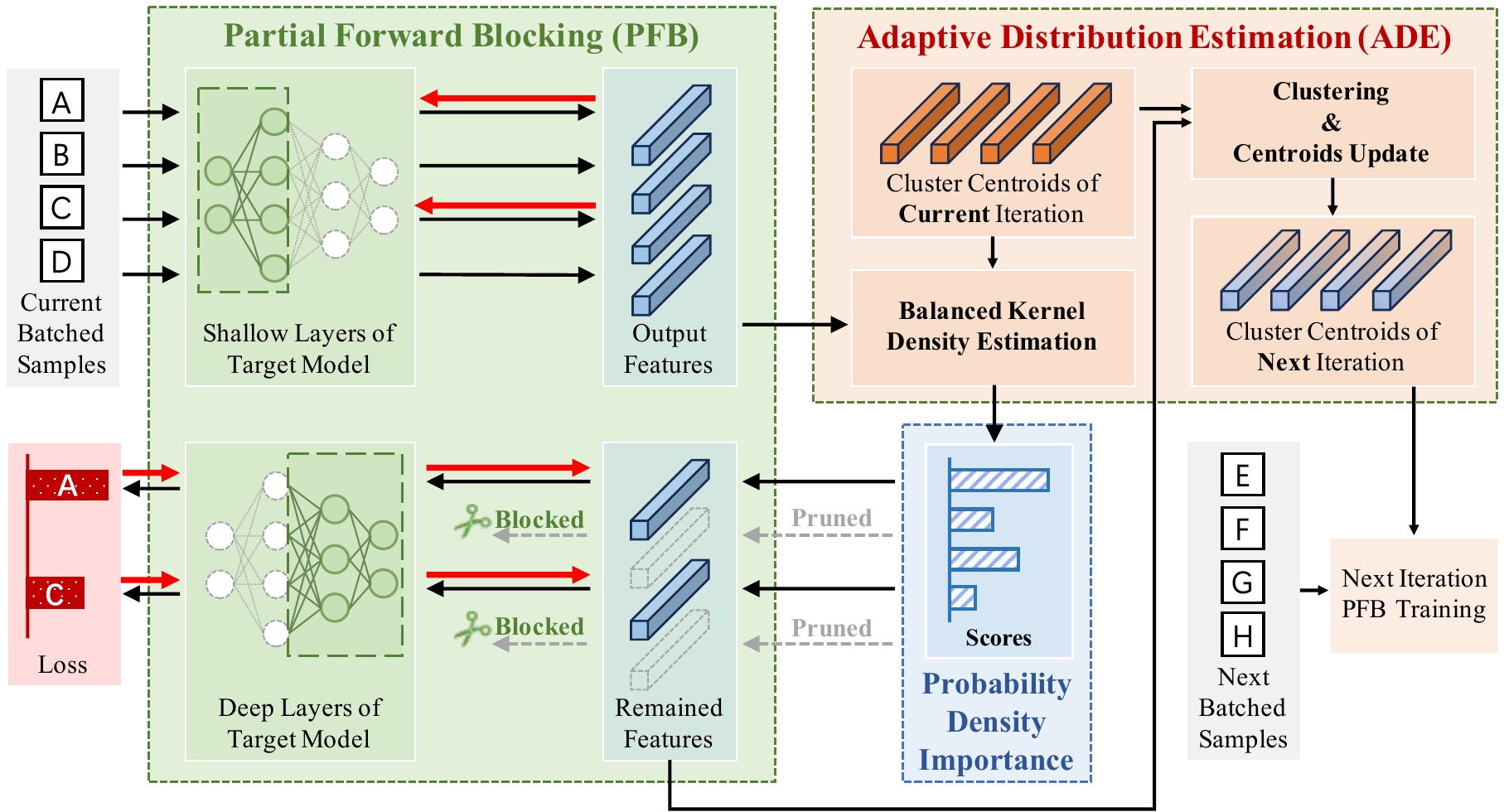}
  \vspace{-3pt}
  \caption{Illustration of the proposed Partial Forward Blocking pipeline. 
  The \textit{black} and \textit{\textcolor{red}{red}} arrows denote forward and backward procedures, respectively. 
  We use A-H to denote individual samples. 
  We adopt an online paradigm which prune samples and save the training cost by block the forward and backward pass of the pruned samples in a batch-wise manner.
  We introduce the Adaptive Distribution Estimation module to obtain probability density, based on which the Probability Density Importance is computed.
  }
   \label{fig:fig2}
\end{figure*}

Recently, dynamic pruning methods that periodically select samples during training have been proposed. 
Raju et al.\cite{raju} first introduced a strategy where the pruning decision for a given sample can change across different pruning cycles. Their proposed methods, UCB and $\epsilon$-greedy, dynamically select different samples for training based on prediction uncertainty at each pruning stage.
Subsequently, Dyn-Unc~\cite{dynunc} adopts a similar pipeline and measures uncertainty in a sliding window of successive training epochs as a pruning metric. 
InfoBatch~\cite{infobatch} mitigates distribution shifts by removing a proportion of low-loss samples while assigning higher weights to the remaining ones in an epoch-wise manner.
Despite their getting rid of the burden of both gradient calculation and proxy model training, they still require a full forward pass on all samples in the dataset for sample-wise importance assessment and pruning.

\subsection{Online Batch Selection}

Online batch selection~\cite{loshchilov2015online, sb} can be regarded as a special type of data pruning, which is more flexible as it selects samples within each batch at every training iteration. Despite this flexibility, many methods still rely on additional surrogate models and expensive gradients, leading to limited efficiency. For example, RHO-Loss~\cite{rho} requires training a surrogate model on a holdout dataset with a certain number of samples, while Deng et al.~\cite{deng2024} use a pre-trained zero-shot predictor as their surrogate model.
On the other hand, while gradient-based online batch selection methods~\cite{katharopoulos2018not, DivBS} save the expense of proxy training, they incur additional computational costs due to the need for loss back-propagation to obtain gradients.

\subsection{Comparison with Existing Methods}

Compared to proxy-model-based methods, our approach eliminates the training cost associated with these models. Unlike gradient-based methods, our method avoids the costly and additional backward propagation for gradient calculation.
Even compared to methods like SB\cite{sb}, UCB\cite{raju} and InfoBatch\cite{infobatch} that mitigate both issues, our approach requires only a small portion of shallow network computations for pruned samples during the forward pass, saving the substantial computational cost in deeper layers. On the contrary, these methods still perform the full, expensive forward pass on pruned samples.

\section{Methodology}

\subsection{Problem Definition}
Given a dataset $\mathit{\Psi}$, our task is to train a deep network $Net_{\theta}$ with parameters $\theta$. 
At the $t$-th iteration during training, a batch of samples $B^t=\{z^t_i\}^{N_B}_{i=1} $ is randomly sampled from $\mathit{\Psi}$.
The samples with the least importance scores in the current $t$-th batch $B^t$ are pruned with a prune ratio $p\in(0,1)$, which can be formulated as optimization problem:
\begin{equation}
\begin{aligned}
  &\mathop{\min}\limits_{m^t_i} \sum_{i=1}^{N_B} m^t_i\mathcal{I}(z^t_i), 
  &\mathit{s.t.} \sum_{i=1}^{N_B}m^t_i=pN_B,
  \label{eq:prune}
\end{aligned}
\end{equation}
where $\mathcal{I}(z^t_i)$ denotes the importance score of $z^t_i$, and $m^t_i \in \{0,1\}$ signifies its pruning decision, where $m^t_i = 1$ indicates $z^t_i$ is pruned.
Hence, by solving \cref{eq:prune} we can get the pruning decisions $M^t=\{m^t_i\}^{N_B}_{i=1}$ and conduct a subset batch $S^t=\{z^t_i | m^t_i=0 \}$ with a smaller size $|S^t|=(1-p)N_B$.
For each training iteration, the target model is only trained on $S^t$. Compared with the vanilla training procedure, the training cost on the subset of the pruned samples $B^t \backslash S^t$ is saved.
Our goal is to maintain strong performance and speed up training simultaneously.

\subsection{Overall Pipeline of Partial Forward Blocking}

The training cost of pruning methods consists of three components: Forward, Backward~\&~Parameter Update (B\&PU), and Scoring. Since most methods perform B\&PU only on retained samples, its cost can not be further reduced.
Thus, \textit{how to effectively reduce the expense of Forward and Scoring} is the key to training acceleration.

Let us begin with analyzing the Forward and Scoring costs of existing methods. Gradient-based approaches rely on loss gradients to compute importance scores, requiring a full forward pass (for Forward) and loss backpropagation (for Scoring) on all samples before pruning. Some methods, such as InfoBatch~\cite{infobatch}, score samples directly based on loss values, reducing Forward and Scoring overhead to a single forward pass on all samples. However, they fail to save the substantial costs of forward pass on pruned samples.
In contrast, proxy-based methods use lightweight models for early pruning, ensuring that only retained samples undergo a forward pass in the target model. However, these approaches require training proxy models for Scoring, which entails a computational cost comparable to training the target model itself, making them impractically expensive.

To overcome these limitations, we propose Partial Forward Blocking (PFB), which computes importance scores and prunes samples in the early stage of forward propagation. Specifically, we derive importance scores from shallow network features, allowing retained samples to reuse these features for the remainder of their forward pass. By pruning early, our method eliminates unnecessary computation on pruned samples in deeper layers, significantly reducing Forward costs compared to gradient-based methods.
Furthermore, for Scoring, we only incur the minor computational cost of shallow layers (e.g., stage-1) without requiring proxy model training. As a result, the total computational cost of Forward and Scoring is limited to a full forward pass on retained samples and a shallow forward pass on pruned ones. This makes our method significantly more efficient than prior approaches. The details are outlined in \cref{alg:ours}. We use $\mathbbm{1}[\cdot]$ to denote the indicator function.

\begin{algorithm}[t!] 
  \caption{Partial Forward Blocking Pipeline}
  \label{alg:ours}
  \SetKwInput{KwIn}{Input}
  \SetKwComment{Comment}{}{}
  \DontPrintSemicolon
  \textbf{Input:}
  Target model $Net(\cdot)$, which can be dived into shallow sub-model $Net^{sh}(\cdot)$ and a deep sub-model $Net^{dp}(\cdot)$, its initial parameters $\theta^0$, pruning ratio $p$. 

  \textbf{Output:}
  The updated parameters $\theta^{T}$. 

  
  \For{$t \in [1,T]$}
  {
  $B^t \leftarrow Sampler(\mathcal{D})$
  
  \Comment*[l]{\textcolor{lightgray}{\small \#Get feature map $\mathbf{f}^t_i \in \mathcal{R}^{ H \times W \times D_{org}}$}}
  $ F^t \leftarrow \{ \mathbf{f}^t_i=Net^{sh}_{\theta^{t\text{-}1}}(z^t_i)~|~z^t_i\in B^t \}$  

  \Comment*[l]{\textcolor{lightgray}{\small \#Pooled representation $\mathbf{x}^t_i \in\mathcal{R}^{1 \times D}$ }}
  $ F^t_{rp} \leftarrow \{ \mathbf{x}^t_i = Pool(\mathbf{f}^t_i)~|~\mathbf{f}^t_i \in F^t \}$\\

  $ I^t \leftarrow \{ \mathcal{I}(z^t_i)=Imp(\mathbf{x}^t_i)~|~ \mathbf{x}^t_i\in F^t_{rp} \}$
  \Comment*[r]{\textcolor{lightgray}{\small \#Eq.2-3}}

  \Comment*[l]{\textcolor{lightgray}{\small \#Prune less important samples }}
  $\tau \leftarrow Percentile(I^t, p)$ 
  \Comment*[r]{\textcolor{lightgray}{\small \#Get threshold}}
  $ M^t \leftarrow \{m^t_i = \mathbbm{1}\left[\mathcal{I}(z^t_i)<\tau \right] ~|~\mathcal{I}(z^t_i) \in I^t \}$ 

  $S^t \leftarrow \{z^t_i | m^t_i=0 \}$

  \Comment*[l]{\textcolor{lightgray}{\small \#Block forward of the pruned ones}}
  $ L^t \leftarrow \{ Loss( Net^{dp}_{\theta^{t\text{-}1}}(\mathbf{f}^t_i))~|~ z^t_i\in S^t \} $ \\
  \vspace{0.1em}
  $ \theta^{t} \leftarrow Update(\theta^{t\text{-}1}, L^t)$

  $t \leftarrow t+1$
  }
  \Return $\theta^{T}$
\end{algorithm}

\subsection{ Probability Density Importance Criterion}
\label{sec:criterion}

It is commonly accepted that strong generalization requires diversified training samples.
\textit{How, then, can we design a per-sample importance criterion to ensure that pruning retains this diversity?}
A natural approach is to eliminate highly redundant samples. If the entire dataset is available, redundancy can be estimated by exhaustively computing pairwise feature similarities.
However, for ultra-large datasets, this brute-force method is impractical due to two major challenges.
First, extracting and storing feature representations for all samples imposes significant storage overhead.
Second, such a similarity calculation demands prohibitive time and computation.

To tackle this problem, we first reframe it in terms of data distribution.
Diversity can be preserved by maintaining a broad distribution and preventing excessive concentration of retained samples.
In the feature distribution, samples in dense regions have many similar counterparts, making them highly redundant and less important.
Even if a significant portion of these samples is removed, their retained counterparts can still preserve the essential information.
Conversely, samples in sparse regions are relatively rare, and removing them would drastically reduce the likelihood of selecting similar samples, losing valuable learning opportunities.
Thus, these samples exhibit low redundancy and high importance.
Therefore, estimating the number or proportion of similar counterparts for each sample provides an indirect measure of its importance.

Following this analysis, we propose leveraging the probability density of a sample's feature distribution as an importance metric. 
Probability density reflects the relative likelihood of a sample with a given feature representation occurring in the dataset.
High probability density suggests a greater prevalence of similar samples, indicating substantial redundancy and diminished importance.
Therefore, for a sample $z_i$, its probability density importance $\mathcal{I}(z_i)$ is defined as follows:
\begin{equation}
  \mathcal{I}(z^t_i) = \frac{1}{f_\mathbf{X}(\mathbf{x}^t_i) + r},
  \label{eq:important}
\end{equation}
where $\mathbf{X}$ denotes the vector-valued random variable~(feature representation) that conforms to the probability distribution of training samples, $f_\mathbf{X}(\cdot)$ is the probability density function (PDF), and $r$ is set as follows:
\begin{equation}
  r=\alpha \cdot \mathop{\max}\limits_{ z_i\in B }f_\mathbf{X}(\mathbf{x}^t_i).
  \label{eq:constant}
\end{equation}
where $\alpha \sim U(0,b)$ is randomly sampled from the uniform distribution, and the upper bound $b$ is set as 0.01.
The ablation study of this random term can be found in \cref{sec:ablation}.
We add the this term~($r$) in \cref{eq:important} 
to complement some randomness of the pruning process 
to further ensure the diversity of training data.

\subsection{Adaptive Distribution Estimation}

In \cref{sec:criterion}, we propose to quantify the importance of all samples in each batch using their probability density.
Nevertheless, \textit{how can we compute the PDF adaptively to track the distribution shift caused by parameter updating while maintaining a negligible computational cost?}

To address this challenge, we propose the Adaptive Distribution Estimation (ADE), which uses Kernel Density Estimation (KDE) to estimate the probability density. By clustering samples in each batch and updating centroids to represent previously retained samples, we reduce the number of kernels in KDE computation. We also track the number of samples each centroid represents and apply weight coefficients to balance kernel contributions. This approach enables efficient, adaptive distribution tracking with minimal computational overhead.

\vspace{0.3em}
\noindent
\textbf{Balanced Kernel Density Estimation.}~
For each sample $z^t_i$ in the current batch $B^t$ of $t$-th training iteration, we conduct spatial average pooling on its original feature map $\mathbf{F}^t_i \in \mathcal{R}^{H\times W\times D_{org}}$ to a $1\times D_{org}$ vector, and then further down-sample on the channel dimension to a $1\times D$ shaped vector $\mathbf{x}^t_i$ as the representation of $z^t_i$, in order to improve the computation efficiency.
Then, kernel functions are applied to the current set of centroids $ C^t = \{ \mathbf{c}^t_j \}^{N_C}_{j=1}$ to estimate the probability density of $z^t_i$ using its representation $\mathbf{x}^t_i$:
\begin{equation}
  \hat{f}_\mathbf{X}(\mathbf{x}^t_i) =  \sum^{N_C}_{j=1} \frac{w^t_j}{N_C} K_{\mathbf{H}}(\mathbf{x}^t_i-\mathbf{c}^t_j),
  \label{eq:kde}
\end{equation}
where $\mathbf{c}^t_j \in \mathcal{R}^{1\times D}$ is the feature of the $j$-th centroid among all $N_C$ ones, and $K_{\mathbf{H}}(\cdot)$ is the scaled kernel function. 
Please note that we introduce $w^t_j$ to balance different kernels,
which will be explained later in \cref{eq:weight}.
For simplicity, we use the standard multivariate normal kernel:
\begin{equation}
  K_{\mathbf{H}}(\mathbf{x})=\frac{1}{ (2\pi)^{d/2} |\mathbf{H}|^{1/2}}~e^{-\frac{1}{2} \mathbf{x}^\top \mathbf{H}^{-1} \mathbf{x}},
  \label{eq:kernel_mg}
\end{equation}
where $\mathbf{H} \in \mathcal{R}^{D\times D}$ is the bandwidth serving as the covariance matrix and can be selected using various bandwidth selection methods~\cite{scott,silverman}.
Following Silverman's rule of thumb\cite{silverman}, $\mathbf{H}$ is set as a diagonal matrix:
\begin{equation}
\begin{aligned}
  \mathbf{H} &= \text{diag}(h_1, h_2, \dots, h_D), \\
  \sqrt{h_d} &= \left( \frac{4}{(D+2)N_C} \right)^\frac{1}{D+4} \sigma_d,
  \label{eq:silver}
\end{aligned}
\end{equation}
where the $\sigma_d$ is the standard deviation of all centroids along the $d$-th dimension.
Thanks to feature map pooling and the small number of centroids, the KDE process is efficient, as evidenced by results in \cref{tab:time}.
Based on \cref{eq:kde,eq:kernel_mg}, we can compute $\mathcal{I}(z^t_i)$ for each $z^t_i$ according to \cref{eq:important}. 
Following the objective in \cref{eq:prune}, we can conduct the small batch $S^t$ consisting of all the retained samples. 

\vspace{0.3em}
\noindent
\textbf{Clustering \& Centroids Update.}

We update centroids and compute weights using only retained samples. Each sample is assigned to its most similar centroid via clustering, formulated as finding a partition $\mathcal{P}(S^t)= \{ A^t_1, A^t_2, \dots, A^t_{N_C} \} $ of $S^t$, where each $z^t_i \in S^t$ belongs to a cluster (subset) $A^t_j$ that minimizes the Euclidean distance to its centroid $\mathbf{c}^t_j$:
\begin{equation}
  \underset{\mathcal{P}(S^t)}{\arg \min} \sum^{N_C}_{j=1} \sum_{z^t_i \in A^t_j} || \mathbf{x}^t_i - \mathbf{c}^t_j ||^2.
  \label{eq:assign}
\end{equation}
Then, centroids are updated for KDE of the next iteration:
\begin{equation}
\begin{aligned}
  \mathbf{c}^{t+1}_j &= \frac{1}{n^{t}_j} \left[ \beta~n^{t-1}_j \cdot \mathbf{c}^{t}_j + (1-\beta) \sum_{z^t_i \in A^t_j} \mathbf{x}^t_i\right], \\
  n^{t}_j &= |A^t_j|+n^{t-1}_j = \sum^t_{iter=1} |A^{iter}_j|,
  \label{eq:update}
\end{aligned}
\end{equation}
where $\beta$=0.01 is the coefficient of EMA, $n^{t-1}_j$ and $n^{t}_j$ are the recorded number of all samples that were previously assigned to the $j$-th centroid until the last and current training iteration, respectively.
These continuously evolving centroids entitle our method with strong adaptability to the shifting distribution.
Moreover, the number of samples previously assigned to each centroid is also used to construct the weight $w^t_j$:
\begin{equation}
  w^{t+1}_j=\frac{n^{t}_j}{t\cdot (1-p)N_B}.
  \label{eq:weight}
\end{equation}
If many similar samples were retained in previous iterations, their assigned centroids gain higher weights, reducing the importance of future similar samples. 
With these weights, our method adaptively favors retaining rarer samples, preserving diversity in the selected subset.

\section{Experiment}
\label{sec:exp}

\subsection{Experimental Setup}

\textbf{Datasets.} Following the most common setting of the latest works\cite{dynunc,DivBS,infobatch}, we conduct experiments to evaluate the performance of PFB on CIFAR-10\cite{cifar}, CIFAR-100\cite{cifar}, and ImageNet-1k\cite{imagenet} for image classification.
CIFAR-10/100 datasets consist of 32$\times$32 sized images that are divided into 10 and 100 classes, respectively. Each of them contains 50,000 images for training and 10,000 for testing. 
ImageNet-1k is a large-scale dataset containing 1,281,167 training images and 50,000 validation images that are categorized into 1,000 classes.
We also evaluate the efficiency of PFB on the semantic segmentation datasets PASCAL VOC 2012 trainaug\cite{voc} and Cityscapes\cite{cityscapes}.
PASCAL VOC 2012 trainaug consists of 10,582 and 1,456 images for training and testing, respectively.
The Cityscapes contains 2,975 fine annotated images with a resolution of 2,048$\times$1,024 for training and another 500 images for validation.

\begin{table}[t!]
  \centering
  \small
  \resizebox{\linewidth}{!}{
  \setlength{\tabcolsep}{4pt}
  \begin{tabular}{ l |c c c |c c c }
      \toprule
       \bf Dataset & \multicolumn{3}{c}{ CIFAR-10} & \multicolumn{3}{|c}{ CIFAR-100}\\
       \midrule
      \bf Pruning Ratio & 30\% & 50\% & 70\% & 30\% & 50\% & 70\% \\
      \midrule
      Random & 94.6 & 93.3 & 90.2 & 73.8 & 72.1 & 69.7 \\

        Herding\cite{herding} & 92.2 & 88.0 & 80.1 & 73.1 & 71.8 & 69.6 \\

        Influence\cite{influence} & 93.1 & 91.3 & 88.3 & 74.4 & 72.0 & 68.9 \\
        
        K-Center\cite{kcenter} & 94.7 & 93.9 & 90.9 & 74.1 & 72.2 & 70.2 \\

        SVP\cite{svp} & 95.0 & 94.5 & 90.3 & 74.2 & 72.3 & 69.8 \\


        Craig\cite{craig} & 94.8 & 93.3 & 88.4 & 74.4 & 71.9 & 69.7 \\
        SB-12hours\textsuperscript{$\dagger$}\cite{sb} & 95.5 & 95.1 & 93.2 & - & - & - \\
        GraNd\cite{EL2N} & 95.3 & 94.6 & 91.2 & 74.6 & 71.4 & 68.8 \\

        Glister\cite{glister} & 95.2 & 94.0 & 90.9 & 74.6 & 73.2 & 70.4 \\
        
        $\epsilon$-greedy\cite{raju} & 95.2 & 94.9 & 94.1 & 76.4 & 74.8 & - \\

       UCB\cite{raju} & 95.3 & 94.7 & 93.9 & 77.3 & 75.3 & - \\

       Forgetting\cite{fogetting} & 94.7 & 94.1 & 91.7 & 75.3 & 73.1 & 69.9 \\

       EL2N\cite{EL2N} & 95.3 & 95.1 & 91.9 & 77.2 & 72.1 & - \\

       Traning Loss\cite{training_loss} & - & - & 94.6 & - & - & 72.6 \\

       AUM\cite{AUM} & 95.1 & 95.3 & 91.4 & 76.9 & 67.4 & 30.6 \\

       Moderate\cite{moderate} & 93.7 & 92.6 & 90.6 & 74.3 & 68.3 & 57.8 \\


       Grad Norm IS\cite{katharopoulos2018not} & - & - & 94.4 & - & - & 73.2 \\

       \textit{Dataset Pruning}\cite{dataset_pruning} & 94.9 & 93.8 & 90.8 & 77.2 & 73.1 & - \\

       CCS\cite{ccs} & 95.4 & 95.0 & 93.0 & 77.1 & 74.5 & 68.9 \\

        MoSo\textsuperscript{$\dagger$}\cite{moso} & - & - & - & 76.7 & 72.3 & 65.8 \\

       InfoBatch*\cite{infobatch} & 95.6 & 95.1 & 94.7 & 78.2 & 78.1 & 76.5 \\


       DivBS\textsuperscript{$\dagger$}\cite{DivBS} & 95.4 & 95.2 & 95.1 & 78.5 & 78.2 & 77.2 \\

\rowcolor[gray]{0.9}
        & \bf 95.9 & \bf 95.5 & \bf 95.2 & \bf 79.1 & \bf 78.8 & \bf 77.9 \\
\rowcolor[gray]{0.9} \multirow{-2}{*}{\bf PFB(Ours)}
        & \textcolor{red}{$\uparrow$0.3} & \textcolor{blue}{$\downarrow$0.1} & \textcolor{blue}{$\downarrow$0.4} & \textcolor{red}{$\uparrow$0.9} & \textcolor{red}{$\uparrow$0.6} & \textcolor{blue}{$\downarrow$0.3} \\

       \midrule
       Full Data & \multicolumn{3}{c|}{95.6\textsubscript{$\pm$0.1}} & \multicolumn{3}{c}{78.2\textsubscript{$\pm$0.1}} \\
       
      \bottomrule
  \end{tabular} 
  }
  \caption{Comparison to state-of-the-art methods on CIFAR-10/100 using ResNet-18. The proposed PFB achieves performance improvements when 30\% samples are pruned. Under 50\% pruning ratio, PFB still achieves nearly lossless results. * denotes different prune ratio settings. $\dagger$ denotes our reproduction. }
  \vspace{-0.5em}
  \label{tab:cifar}
\end{table} 

\begin{table}[t!]
  \centering
  \begin{tabular}{ l c | c c c }
      \toprule
       \multirow{11}{*}{ \rotatebox[origin=c]{90}{\shortstack{ResNet-50}}} & \bf Pruning Ratio  & 30\% & 50\% & 70\% \\
      \cmidrule{2-5}
       
       & Random & 72.2\textsubscript{\textcolor{blue}{$\downarrow$4.2}} & 69.1\textsubscript{\textcolor{blue}{$\downarrow$7.3}} & 65.9\textsubscript{\textcolor{blue}{$\downarrow$10.5}}  \\

       & Herding\cite{herding} & 73.5\textsubscript{\textcolor{blue}{$\downarrow$2.9}} & 69.3\textsubscript{\textcolor{blue}{$\downarrow$7.1}} & 65.1\textsubscript{\textcolor{blue}{$\downarrow$11.3}}  \\

       & Forgetting\cite{fogetting} & 74.8\textsubscript{\textcolor{blue}{$\downarrow$1.6}} & 72.0\textsubscript{\textcolor{blue}{$\downarrow$4.4}} & 67.8\textsubscript{\textcolor{blue}{$\downarrow$8.6}}\\

       & EL2N\cite{EL2N} & 74.3\textsubscript{\textcolor{blue}{$\downarrow$2.1}} & 68.5\textsubscript{\textcolor{blue}{$\downarrow$7.9}} & 54.8\textsubscript{\textcolor{blue}{$\downarrow$21.6}} \\

       & Moderate\cite{moderate} & 75.2\textsubscript{\textcolor{blue}{$\downarrow$1.2}} & 72.2\textsubscript{\textcolor{blue}{$\downarrow$4.2}} & 67.7\textsubscript{\textcolor{blue}{$\downarrow$8.7}} \\

       & MoSo\textsuperscript{$\dagger$}\cite{moso} & 76.5\textsubscript{\textcolor{red}{$\uparrow$0.1}} & 73.5\textsubscript{\textcolor{blue}{$\downarrow$2.9}} & 70.0\textsubscript{\textcolor{blue}{$\downarrow$6.4}} \\

       & InfoBatch\textsuperscript{$\dagger$}\cite{infobatch} & 76.5\textsubscript{\textcolor{red}{$\uparrow$0.1}} & 75.8\textsubscript{\textcolor{blue}{$\downarrow$0.6}} & 74.9\textsubscript{\textcolor{blue}{$\downarrow$1.5}} \\
\rowcolor[gray]{0.9}
       \cellcolor{white} & \bf PFB(Ours) & \bf 77.0\textsubscript{\textcolor{red}{$\uparrow$0.6}} & \bf 76.1\textsubscript{\textcolor{blue}{$\downarrow$0.3}} & \bf 75.3\textsubscript{\textcolor{blue}{$\downarrow$1.1}} \\
       
       \cmidrule{2-5}
       & Full Data & \multicolumn{3}{c}{76.4\textsubscript{$\pm$0.2}} \\
        \bottomrule
        \toprule
       \multirow{11}{*}{ \rotatebox[origin=c]{90}{\shortstack{Swin-T}}} & \bf Pruning Ratio & 30\% & 40\% & 50\% \\
        \cmidrule{2-5}
        & Random & 77.2\textsubscript{\textcolor{blue}{$\downarrow$2.4}} & 75.9\textsubscript{\textcolor{blue}{$\downarrow$3.7}} & 74.5\textsubscript{\textcolor{blue}{$\downarrow$5.1}}  \\
        & Forgetting\cite{fogetting} & 78.3\textsubscript{\textcolor{blue}{$\downarrow$1.3}} & 77.6\textsubscript{\textcolor{blue}{$\downarrow$2.0}} & 74.3\textsubscript{\textcolor{blue}{$\downarrow$5.3}}\\
        & EL2N\cite{EL2N} & 78.2\textsubscript{\textcolor{blue}{$\downarrow$1.4}} & 75.9\textsubscript{\textcolor{blue}{$\downarrow$3.7}} & 71.1\textsubscript{\textcolor{blue}{$\downarrow$8.5}} \\
        & SVP\cite{svp} & 76.6\textsubscript{\textcolor{blue}{$\downarrow$3.0}} & 74.9\textsubscript{\textcolor{blue}{$\downarrow$4.7}} & 72.7\textsubscript{\textcolor{blue}{$\downarrow$6.9}} \\
        & Moderate\cite{moderate} & 77.1\textsubscript{\textcolor{blue}{$\downarrow$2.5}} & 75.9\textsubscript{\textcolor{blue}{$\downarrow$3.7}} & 75.0\textsubscript{\textcolor{blue}{$\downarrow$4.6}} \\
        & InfoBatch\textsuperscript{$\dagger$}\cite{moderate} & 78.6\textsubscript{\textcolor{blue}{$\downarrow$1.0}} & 78.2\textsubscript{\textcolor{blue}{$\downarrow$1.4}} & 77.5\textsubscript{\textcolor{blue}{$\downarrow$2.1}} \\
        & Dyn-Unc\cite{dynunc} & 79.1\textsubscript{\textcolor{blue}{$\downarrow$0.5}} & 78.5\textsubscript{\textcolor{blue}{$\downarrow$1.1}} & 77.6\textsubscript{\textcolor{blue}{$\downarrow$2.0}} \\
\rowcolor[gray]{0.9}
       \cellcolor{white} & \bf PFB(Ours) & \bf 79.6\textsubscript{\textcolor{red}{$\uparrow$0.0}} & \bf 79.2\textsubscript{\textcolor{blue}{$\downarrow$0.4}} & \bf 78.2\textsubscript{\textcolor{blue}{$\downarrow$1.4}} \\
        \cmidrule{2-5}
       & Full Data & \multicolumn{3}{c}{79.6\textsubscript{$\pm$0.1}} \\
      \bottomrule
  \end{tabular} 
  \caption{Performance Comparison on ImageNet-1k. ResNet-50 and Swin-T are trained from scratch for 90 and 300 epochs, respectively.
  The results demonstrate the effectiveness of PFB on both CNNs and Transformers.}
  \vspace{-0.6em}
  \label{tab:imagenet}
\end{table}

\noindent
\textbf{Implementation details.}
For image classification, we employ ResNet-18\cite{resnet} as a backbone on CIFAR-10/100 trained for 200 epochs.
The batch size is set as 128.
On ImageNet-1k, we evaluate the performance of PFB using ResNet-50 and Swin-T\cite{liu2021swin} to evaluate the generalization across different network architectures.
The batch size of both ResNet-50 and Swin-T is set as 1024.
Other detailed settings can be found in our supplementary materials.

\begin{table*}[t!]
  \centering
  \small
    \setlength{\tabcolsep}{4pt}
  \begin{tabular}{ l c c | c | c c c c }
        \toprule
        \bf Method & \bf Year & \bf Pruning Freq. & \bf Top-1 Acc(\%) & \bf Training(h) & \bf Overhead(h) & \bf Total(n*h) & \bf Reduction \\
        \midrule
        Full Data & - & - & 76.4 & 13.9 & - & 55.6 & - \\
        \midrule  
        EL2N\textsuperscript{$\dagger$}\cite{EL2N} & 2018 & Single-shot & 71.5\textsubscript{\textcolor{blue}{$\downarrow$4.9} } & 10.1 & \textgreater14 & \textgreater96 & \textgreater72.7\%$\uparrow$  \\
        UCB\textsuperscript{$\dagger$}\cite{raju} & 2021 & Epoch & 75.8\textsubscript{\textcolor{blue}{$\downarrow$0.6} }  & 10.1 & 0.08 & 40.8 & 26.7\%$\downarrow$ \\
        InfoBatch\textsuperscript{$\dagger$}\cite{infobatch}& 2023 & Epoch & 76.5\textsubscript{\textcolor{red}{$\uparrow$0.1} }  & 10.1 & 0.07 & 40.7 & 26.8\%$\downarrow$ \\
        DivBS\textsuperscript{$\dagger$}\cite{DivBS} & 2024 & Batch & 76.4\textsubscript{\textcolor{red}{$\uparrow$0.0} }  & 11.2 & 0.72 & 47.6 & 14.4\%$\downarrow$ \\
\rowcolor[gray]{0.9}
        \bf PFB(ours) & - & Batch & \bf 76.9\textsubscript{\textcolor{red}{$\uparrow$0.5} }  & \bf 9.2 & \bf 0.06 & \bf 37.1 & \bf 33.2\%$\downarrow$ \\
        
        \bottomrule
  \end{tabular} 
  \caption{Time cost comparison on ImageNet-1k using ResNet-50 under 40\% pruning ratio. PFB saves the overall cost by more than 30\% but still obtains a 0.5\% accuracy gain. Results are collected on a 4-RTX 4090 GPU server. Automatic Mixed Precision\cite{micikevicius2018mixed} is adopted during training for all methods. `Training' and `Overheads' denote the wall clock time spent on network training and the additional time brought in by pruning methods. `Total(n*h)' is the total node hour with n=4. }
  \vspace{-0.5em}
  \label{tab:time}
\end{table*} 

\begin{table}[t!]
  \centering
  \small
  \begin{tabular}{ c l c c c }
      \toprule
       \bf Dataset & \bf Method & \bf mIoU & \bf \shortstack{Time Saved} \\
        \midrule
        \multirow{4}{*}{ \shortstack{Cityscapes\\ 30\% Pruned}  }& Full Data  & 80.4 & - \\
        \cmidrule{2-4}
        & DivBS\textsuperscript{$\dagger$} & 79.0\textsubscript{\textcolor{blue}{$\downarrow$1.4}} & 7.8\%$\downarrow$ \\
        & InfoBatch\textsuperscript{$\dagger$} & 79.2\textsubscript{\textcolor{blue}{$\downarrow$1.2}} & 15.8\%$\downarrow$ \\
\rowcolor[gray]{0.9}
        \cellcolor{white} & \bf PFB(Ours) & \bf 79.8\textsubscript{\textcolor{blue}{$\downarrow$0.6}} & \bf 24.1\%$\downarrow$ \\
      \midrule
        \multirow{4}{*}{ \shortstack{VOC 2012\\ 70\% Pruned}  }& Full Data  & 77.9 & - \\
        \cmidrule{2-4}
        & DivBS\textsuperscript{$\dagger$} & 72.7\textsubscript{\textcolor{blue}{$\downarrow$5.2}} & 42.6\%$\downarrow$ \\
        & InfoBatch\textsuperscript{$\dagger$} & 72.4\textsubscript{\textcolor{blue}{$\downarrow$5.5}} & 49.1\%$\downarrow$ \\
\rowcolor[gray]{0.9}
        \cellcolor{white} & \bf PFB(Ours) & \bf 73.4\textsubscript{\textcolor{blue}{$\downarrow$4.5}} & \bf 56.7\%$\downarrow$ \\
      \bottomrule
  \end{tabular} 
  \caption{Results comparison with other pruning methods on Cityscapes and Pascal VOC 2012. PFB blocks the forward pass of pruned images at the third stage of the encoder. All methods adopt the same training settings.}
  \vspace{-0.5em}
  \label{tab:seg}
\end{table}

\subsection{Comparisons with SOTA Methods}
We compare our method with existing state-of-the-art data pruning and online batch selection methods on CIFAR-10/100 and ImageNet-1k.
We define the pruning $p$ as the proportion of pruned samples to all samples.
For methods that prune the whole training dataset, like most data pruning methods\cite{EL2N,infobatch}, the pruning ratio is set as $p=\frac{|\mathcal{D}|}{|\mathcal{D}_{pruned}|}$.
As for methods (including ours) that adopt batch-wise pruning paradigm\cite{DivBS,sb}, the pruning ratio is set as $p=\frac{N_S}{N_B}$.
Note that InfoBatch\cite{infobatch} adopts a different pruning rate calculation strategy: it first selects only part of the samples with the lower loss as well-learned and applies pruning only within this subset, resulting in an actual pruning rate that is around only half of the reported value. 
To distinguish this, we denote the original InfoBatch with `*' and our reproduction with `$\dagger$', which considers 95\% of the samples as pruning candidates.
We also introduce single-shot random pruning~(Random) as a baseline for comparison. 
Other reproduced results following the same experimental setup of PFB are also denoted with `$\dagger$'.

\vspace{0.3em}
\noindent
\textbf{Performance Comparisons.}
Top-1 accuracy on CIFAR-10/100 and ImageNet-1K is reported in \cref{tab:cifar} and \cref{tab:imagenet}, respectively.
The results demonstrate that PFB consistently outperforms existing methods across all three datasets and various pruning ratios.
Notably, at a 30\% pruning ratio, PFB achieves better-than-lossless training, with ResNet models improving by 0.3\%, 0.9\%, and 0.6\% over full-data training on CIFAR-10, CIFAR-100, and ImageNet-1K, respectively.
These results strongly validate the diversity of the retained training samples and the enhanced generalization of the trained models.
We attribute this improvement to both the proposed importance criterion and the Adaptive Distribution Estimation module.

\smallskip
\noindent
\textbf{Efficiency Comparisons on ImageNet-1k.}
The training efficiency of PFB is also compared with both classic methods~(EL2N and UCB) and the latest methods~(InfoBatch and DivBS).
The time cost of ResNet-50 trained on ImageNet-1k for 90 epochs is reported in \cref{tab:time}. 
EL2N and DivBS are representative proxy-based and gradient-based methods, respectively. As a result, they struggle to achieve satisfactory speedup on large-scale datasets like ImageNet due to their inherent computational overhead. In contrast, UCB and InfoBatch rank sample importance based on prediction uncertainty and loss values, effectively mitigating the high cost associated with proxy models and backpropagation in the previous approaches. Despite this, our proposed PFB mechanism further reduces training overhead by additionally skipping part of the forward pass for pruned samples, leading to even greater efficiency gains.

\noindent
\textbf{Efficiency Comparisons on Segmentation Datasets.}
To better demonstrate the acceleration advantage of PFB in skipping part of the forward pass, we further conduct experiments on a more computationally intensive semantic segmentation task, with results presented in \cref{tab:seg}. We apply InfoBatch, DivBS, and PFB to DeepLabV3-ResNet50\cite{deeplabv3}, ensuring all methods undergo the same number of training iterations. 
Since semantic segmentation is a dense prediction task that requires per-pixel supervision, its forward inference and loss computation demand significantly more computational resources and time compared to classification tasks. 
Consequently, PFB achieves a more pronounced efficiency gain over InfoBatch and DivBS by skipping the forward pass through deep layers. 
Furthermore, as PFB does not rely on gradient-based importance estimation, it exhibits a greater advantage over DivBS on Cityscapes than VOC, where images are larger and computational demands are higher. 
Experimental results show that with the PFB mechanism, PFB achieves a threefold reduction in training time compared to DivBS on Cityscapes.

\begin{figure*}[t!]
  \centering
  \setlength{\tabcolsep}{2pt}
  \begin{tabular}{ cccc }
  \includegraphics[width=0.3\linewidth]{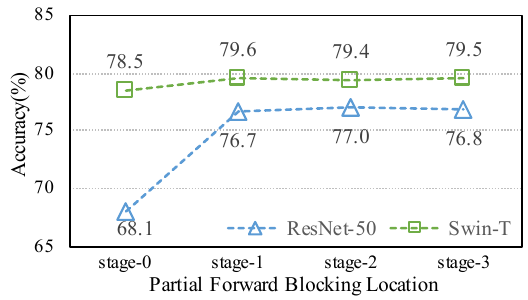}
  & 
  \includegraphics[width=0.3\linewidth]{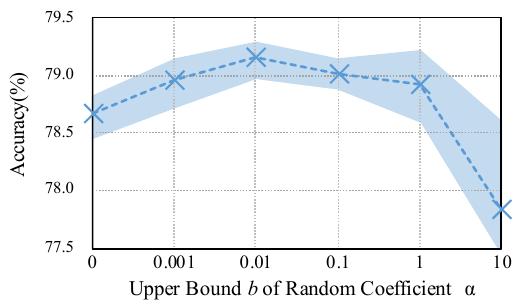} 
  &
  \includegraphics[width=0.3\linewidth]{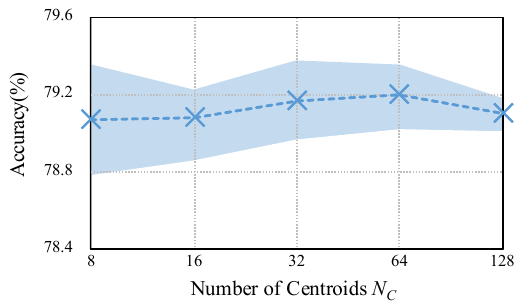} 
  \\
  \vspace{-3pt}
  \small 
  (a) Location to Block & \small (b) Scale of Random Term & \small (c) Number of Centroids \\
  
  \includegraphics[width=0.3\linewidth]{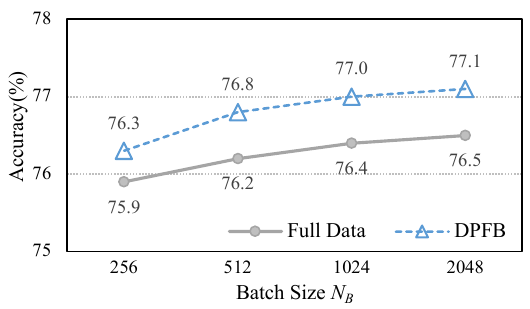}
  &
  \includegraphics[width=0.3\linewidth]{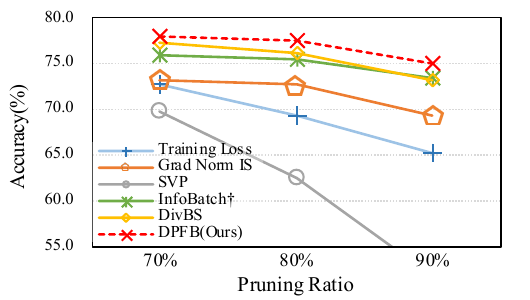}
  &
  \includegraphics[width=0.3\linewidth]{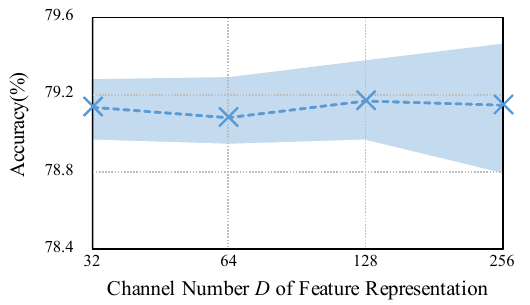} \\
  \small 
  (d) Different Batch Size & \small (e) Large Pruning Ratios & \small (f) Representation Dimension 
  \vspace{-5pt}
  \end{tabular} 
    \caption{Ablation studies of PFB on (a) the blocking location of PFB strategy on ImageNet-1k, (b) the upper bound of the uniformly sampled coefficient of the random term, (c) the number of centroids for KDE, (d) the size of batch on ImageNet-1k, (e) performance under large pruning ratios, and (f) the channel number of feature representation. We use the shading to show the maximum and minimum of five independent runs. Unless otherwise specified, experiments are conducted on CIFAR-100.
    \vspace{-0.5em}
  }
   \label{fig:ab}
\end{figure*}

\subsection{Ablation Study}
\label{sec:ablation}

\noindent
\textbf{Location to Block.}
We first investigate the impact of the blocking location on performance.
The results in \cref{fig:ab}(a) show that For both ResNet-50 and Swin-T, PFB performs well when blocking after stage-1 layers.
When PFB is applied directly at the stem layers (stage-0) of ResNet-50, using the corresponding features as sample representations, we observe a significant drop in performance. 
We conjecture that this is due to the relatively noisy and less informative nature of shallow features in ResNet, which may not sufficiently capture the semantic distinctions and relationships between samples.
To balance performance and efficiency, we therefore apply PFB at stage-2 for ResNet and stage-1 for Swin.

\vspace{0.2em}
\noindent
\textbf{Batch Size.}
As the recorded centroids and the number of samples assigned to it is related to the number of samples in the batch, we conduct experiments to verify the robustness of PFB to different batch size settings on ImageNet-1k.
Results in \cref{fig:ab}(d) reveal that the trend of PFB is consistent with that of full-data training.
This phenomenon demonstrates that PFB is not sensitive to various batch sizes.

\vspace{0.1em}
\noindent
\textbf{Random Term of Importance.}
The random term $r$ in our proposed importance metric (\cref{eq:important}) enhances the diversity of retained samples, ensuring those with plenty of similar counterparts can sometimes be retained as well.
We conducted five repeated experiments on the upper bound $b$, which controls the scale of $r$ on CIFAR-100.
We report the results in \cref{fig:ab}(b) with blue shading indicating the range between the maximum and minimum values.
The results demonstrate that the accuracy remains stable when the upper bound $b$ of random coefficient $\alpha$ is within [0.001, 1]. 
However, when $\alpha$ is excessively large, the random term dominates probability density in (\cref{eq:important}), leading to an excessive retention of randomly selected samples.
Thus, the performance drops significantly with large fluctuations when $b$ is set as 10.

\vspace{0.1em}
\noindent
\textbf{Effectiveness on High Pruning Ratio.}
To explore the effectiveness under extreme conditions, we conduct ablation studies on CIFAR-100 with very large pruning ratios(0.7-0.9).
As shown in \cref{fig:ab}(e), PFB still outperforms other powerful methods, including both gradient-based ones\cite{DivBS,katharopoulos2018not} and those relying on proxy-models\cite{svp}.

\vspace{0.1em}
\noindent
\textbf{Coefficients in Adaptive Distribution Estimation.}
To find the best coefficients for probability density estimation, we compare the accuracy and the fluctuation under different settings in \cref{fig:ab}(c) and (f). 
The results show that our method is robust to variations in the number of centroids $N_{C}$ and feature dimensions $D$ within a normal range. Moreover, a larger $N_{C}$ and a lower $D$ help to reduce performance fluctuations. 
This aligns with intuition: increasing the kernel number within a reasonable range improves KDE estimation accuracy, while a higher dimension increases the complexity and difficulty. We set $N_C$=64 and $D$=128.

\vspace{-0.1em}
\section{Conclusion}
\label{sec:conclusion}
In this paper, we introduce Partial Forward Blocking (PFB), a novel training data compression paradigm designed to accelerate deep model training. Unlike existing dataset pruning methods that rely on computationally expensive gradient-based importance evaluation or additional proxy models, PFB efficiently selects samples by leveraging a lightweight probability density estimation mechanism without compromising model generalization. Furthermore, we propose a Partial Forward Blocking (PFB) strategy, which skips deep-layer computations for pruned samples, significantly reducing training costs. 
Extensive experiments demonstrate that PFB achieves superior training efficiency while outperforming state-of-the-art data pruning and online batch selection methods on both classification and segmentation datasets. Our approach provides a scalable and adaptive solution for large-scale dataset training, paving the way for more efficient deep learning models.

\section{Acknowledgement}

This work was supported by the National Natural Science Foundation of China No.62176097, and the Hubei Provincial Natural Science Foundation of China No.2022CFA055.

{
    \small
    \bibliographystyle{ieeenat_fullname}
    \bibliography{main}
}
\appendix

\section{More Experimental Results}

In this section, we provide more experimental results to help
our readers better understand our proposed method.

\subsection{Ablation Study on the Weight in ADE}

To enhance the adaptability of PFB to the changing distribution caused by parameter updating, we introduce the $w^t_j$ (Eq.~(4) and Eq.~(9)) as a dynamic weight to adaptively balance between different centroids.
By re-scaling the output of each kernel according to the number of samples that were previously assigned to its cluster (represented by the centroids), those centroids with a great number of $n^t_j$ will contribute more to the probability density, thereby reducing the importance score of samples with similar features.
We evaluate its effectiveness by replacing the $w^t_j$ with 1 (denoted as `w/o weight') and comparing the results with PFB in \cref{tab:appendix_weight}.
Although PFB w/o weight still outperforms DivBS and InfoBatch, the performance drops by a large margin when the prune ratio grows.
This phenomenon shows the effectiveness of the balancing weight, which may lead to a better estimation of the probability density function.

\begin{table}[h!]
  \centering
    \begin{tabular}{c|ccc}
    \toprule
    \bf Pruning Ratio & 30\% & 50\% & 70\% \\
    \midrule
    Full Data & \multicolumn{3}{c}{78.2} \\
    \midrule
    InfoBatch*\cite{infobatch} & 78.2\textsubscript{\textcolor{red}{$\uparrow$0.0}} & 78.1\textsubscript{\textcolor{blue}{$\downarrow$0.1}} & 76.5\textsubscript{\textcolor{blue}{$\downarrow$1.7}}  \\
    DivBS\cite{DivBS} & 78.5\textsubscript{\textcolor{red}{$\uparrow$0.3}} & 78.2\textsubscript{\textcolor{red}{$\uparrow$0.0}} & 77.2\textsubscript{\textcolor{blue}{$\downarrow$1.0}} \\
    PFB w/o weight & 78.9\textsubscript{\textcolor{red}{$\uparrow$0.7}} & 78.4\textsubscript{\textcolor{red}{$\uparrow$0.2}} & 77.4\textsubscript{\textcolor{blue}{$\downarrow$0.8}} \\
    \bf PFB(ours) & \bf 79.1\textsubscript{\textcolor{red}{$\uparrow$0.9}} & \bf 78.8\textsubscript{\textcolor{red}{$\uparrow$0.6}} & \bf 77.9\textsubscript{\textcolor{blue}{$\downarrow$0.3}} \\
    \bottomrule
    \end{tabular}
\caption{Ablation study on the weight in ADE for balancing different centroids. Experiments are conducted on CIFAR-100 using ResNet-18. We use `w/o weight' to denote our modification that $w^t_j$ is replaced by 1 in Eq.~(4) of the main text.}
\label{tab:appendix_weight}
\end{table}

\subsection{Ablation Study on the Bandwidth Estimation Methods in ADE}

\begin{table}[h!]
  \small
  \centering
  \resizebox{\linewidth}{!}{
    \begin{tabular}{c|c|ccc}
    \toprule
    \bf Methods & \bf Bandwidth & 30\% & 50\% & 70\% \\
    \midrule
    Full Data  & - & \multicolumn{3}{c}{78.2} \\
    \midrule
    InfoBatch*\cite{infobatch} & - & 78.2\textsubscript{\textcolor{red}{$\uparrow$0.0}}  & 78.1\textsubscript{\textcolor{blue}{$\downarrow$0.1}} & 76.5\textsubscript{\textcolor{blue}{$\downarrow$1.7}}  \\
    DivBS\cite{DivBS} & - & 78.5\textsubscript{\textcolor{red}{$\uparrow$0.3}} & 78.2\textsubscript{\textcolor{red}{$\uparrow$0.0}} & 77.2\textsubscript{\textcolor{blue}{$\downarrow$1.0}} \\
    \midrule
    \multirow{3}{*}{PFB} & Identity & 77.4\textsubscript{\textcolor{blue}{$\downarrow$0.8}}  & 76.5\textsubscript{\textcolor{blue}{$\downarrow$1.7}} & 75.1\textsubscript{\textcolor{blue}{$\downarrow$3.1}} \\
     & Scott\cite{scott}  & 79.0\textsubscript{\textcolor{red}{$\uparrow$0.8}} & \bf 78.8\textsubscript{\textcolor{red}{$\uparrow$0.6}} & \bf 77.9\textsubscript{\textcolor{blue}{$\downarrow$0.3}} \\
     & Silverman\cite{silverman} & \bf 79.1\textsubscript{\textcolor{red}{$\uparrow$0.9}} & \bf 78.8\textsubscript{\textcolor{red}{$\uparrow$0.6}} & \bf 77.9\textsubscript{\textcolor{blue}{$\downarrow$0.3}} \\
    \bottomrule
    \end{tabular}
    }
\caption{Ablation study on different bandwidth estimation methods. Experiments are conducted on CIFAR-100 using ResNet-18.}
\label{tab:appendix_bandwidth}
\end{table}

As bandwidth estimation is usually deemed important for KDE methods, we compare the performance of two popular bandwidth estimation rules, namely Scott's rule\cite{scott} and Silverman's rule\cite{silverman}.
We also introduce a baseline denoted as `Identity'.
This baseline simply sets the $\mathbf{H}$ as an identity matrix.
Results in \cref{tab:appendix_bandwidth} indicate that a proper bandwidth is crucial for the performance of our PFB. 
However, there is no big difference between those two commonly used bandwidth estimation methods.

\subsection{Detailed Explanation of InfoBatch}

InfoBatch\cite{infobatch} employs a soft pruning ratio, using the mean loss value of all samples as a threshold to divide the dataset into two subsets.
Samples with a loss lower than this threshold form a candidate subset, where each sample has a pruning probability of $p$, which is reported as the pruning ratio.
However, since the candidate subset only contains half of the samples, the actual pruning ratio of InfoBatch is $p/2$.
To highlight this distinction, we denote the original version of InfoBatch with `*' in the main text.
For a fair comparison, we follow the modification applied by DivBS\cite{DivBS} to InfoBatch, where the threshold is set to the 95\% percentile to align with the actual pruning ratio of most pruning methods.
We denote this modified version as InfoBatch\textsuperscript{$\dagger$}.
In the main text, Tab.~2-4 present a comparison between our method and InfoBatch\textsuperscript{$\dagger$} on ImageNet-1k, Cityscapes, and PASCAL VOC 2012.
Here, we further provide experimental results on CIFAR-100 in \cref{tab:appendix_infobatch} to supplement the comparison.
Aligning the actual pruning ratio reveals a significant performance drop for InfoBatch on CIFAR-100.
Beyond the impact of the adjusted pruning ratio, this decline may also stem from the large weights applied to the retained samples within the pruned candidate subset by InfoBatch. (Please refer to \cite{infobatch} for details of this re-scaling operation.)
Such a large weight may excessively emphasize the retained samples, potentially hindering the learning of harder examples in the other subset.

\begin{table}[h!]
  \centering
    \begin{tabular}{cccc}
    \toprule
     \textcolor{gray}{Pruning Ratio} & \textcolor{gray}{30\%} & \textcolor{gray}{50\%} & \textcolor{gray}{70\%} \\
    \bf Actual Ratio & 15\% & 25\% & 35\% \\
    \midrule
    InfoBatch*\cite{infobatch} & 78.2\textsubscript{\textcolor{red}{$\uparrow$0.0}} & 78.1\textsubscript{\textcolor{blue}{$\downarrow$0.1}} & 76.5\textsubscript{\textcolor{blue}{$\downarrow$1.7}}  \\
    \midrule
    \midrule
    \bf Pruning Ratio & 30\% & 50\% & 70\% \\
    \midrule
    InfoBatch\textsuperscript{$\dagger$} & 78.0\textsubscript{\textcolor{blue}{$\downarrow$0.2}} & 76.0\textsubscript{\textcolor{blue}{$\downarrow$2.2}} & 74.3\textsubscript{\textcolor{blue}{$\downarrow$3.9}}  \\
    DivBS\cite{DivBS} & 78.5\textsubscript{\textcolor{red}{$\uparrow$0.3}} & 78.2\textsubscript{\textcolor{red}{$\uparrow$0.0}} & 77.2\textsubscript{\textcolor{blue}{$\downarrow$1.0}} \\
    \bf PFB(ours) & \bf 79.1\textsubscript{\textcolor{red}{$\uparrow$0.9}} & \bf 78.8\textsubscript{\textcolor{red}{$\uparrow$0.6}} & \bf 77.9\textsubscript{\textcolor{blue}{$\downarrow$0.3}} \\
    \midrule
    Full Data & \multicolumn{3}{c}{78.2} \\
    \bottomrule
    \end{tabular}
\caption{Fair comparison on CIFAR-100 with InfoBatch.}
\label{tab:appendix_infobatch}
\end{table} 

\subsection{Error Bars}

Error statistics for ResNet-18 and Swin-T on different datasets are also included in \cref{tab:error}, exhibiting variations within an acceptable range. 
The results show that PFB can maintain a stable performance with negligible variation.

\begin{table}[h!]
  \centering
  \tiny
  \resizebox{0.99\linewidth}{!}{
    \setlength{\tabcolsep}{2pt}
  \begin{tabular}{ l | c c c | c c c | c c c }
      \hline
       \bf Dataset/Model & \multicolumn{3}{c|}{ CIFAR-10~/~ResNet-18} & \multicolumn{3}{c|}{ CIFAR-100~/~ResNet-18} & \multicolumn{3}{c}{ ImageNet-1K~/~Swin-T} \\
       \hline
      \bf Pruning Ratio & 30\% & 50\% & 70\% & 30\% & 50\% & 70\% & 30\% & 40\% & 50\% \\
      \hline

\rowcolor[gray]{0.9}
        & 95.9 & 95.5 & 95.2 & 79.1 & 78.8 & 77.9 & 79.6 & 79.2 & 78.2\\
\rowcolor[gray]{0.9} \multirow{-2}{*}{\bf PFB(Ours)}
        &$\pm$0.1 &$\pm$0.1&$\pm$0.2
        &$\pm$0.2 &$\pm$0.2&$\pm$0.2
        &$\pm$0.1 &$\pm$0.2&$\pm$0.2\\

       \hline
       Full Data & \multicolumn{3}{c|}{95.6~$\pm$0.1}& \multicolumn{3}{c|}{78.2~$\pm$0.1} & \multicolumn{3}{c}{79.6~$\pm$0.1} \\
       
      \hline
  \end{tabular} 
  }
  \caption{Error bars on CIFAR-10/100 and ImageNet-1k.}
  \label{tab:error}
\end{table}

\section{Implementation Details}

We further demonstrate the details of experiments on image classification and segmentation datasets here.

\subsection{Classification Training Settings}
All the classification experiments are conducted on a 4-RTX 4090 GPU server.
We follow the training details of InfoBatch\cite{infobatch} on CIFAR-10/100 using ResNet-18 and ImageNet-1k using ResNet-50.
For Swin-T, we adopt similar training settings of Dyn-Unc\cite{dynunc}.
The AutoAugment\cite{cubuk2018autoaugment} is applied to augment training data only for Swin-T, including random path drop and gradient clipping for a fair comparison with Dyn-Unc\cite{dynunc} in Tab.~(2) of the main text.
All the detailed settings needed for reproduction are listed in the \cref{tab:appendix_details}.

\subsection{Details of Segmentation Experiments}
Our segmentation experiments are based on the implementation of MMSegmentation\cite{mmseg}. 
On PASCAL VOC 2012\cite{voc} and Cityscapes\cite{cityscapes}, the models are trained for 36,000 iterations. 
Other training and evaluation details remain the same with MMSegmentation.
Please note that the reported mIoU results in Tab.~(3) and (4) employ the popular multi-scale evaluation technique.
Moreover, as most semantic segmentation methods employ an auxiliary segmentation head at the third stage of the encoder, we extract the features at this stage to utilize the abundant semantic information.
Please note that there is a big difference between segmentation and classification networks: aside from the encoder, segmentation networks usually have a computationally expensive decoder. 
Hence, blocking at the third stage of the encoder can still significantly cut down the training time.

\begin{table}[h!]
  \centering
  \resizebox{\linewidth}{!}{
    \begin{tabular}{cl|c|c|cc}
    \toprule
     & \textbf{Parameters}  & \bf CIFAR-10 & \bf CIFAR-100 & \multicolumn{2}{c}{\bf ImageNet-1k} \\
     \midrule
     & \bf Models & ResNet-18 & ResNet-18 & ResNet-50 & Swin-T \\
    \midrule  
    \multirow{8}{*}{\bf \rotatebox[origin=c]{90}{Training}} 
    & optimizer & SGD & SGD & Lars & AdamW \\
    & weight\_decay & 0.0005 & 0.0005 & 0.00005 & 0.05 \\
    & batch\_size & 128 & 128 & 1024 & 1024 \\
    & epochs & 200 & 200 & 90 & 300 \\
    & learning\_rate & 0.10 &  0.05  & 6.4 & 0.001  \\
    & label smoothing & 0.1 & 0.1 & 0.1 & 0.1 \\
    & learning rate scheduler & OncCycle & OncCycle & OneCycle & CosineAnnealing \\
    & learning rate warmup & - & - & 5 & 20 \\
    \midrule 
    \multirow{6}{*}{\bf \rotatebox[origin=c]{90}{Data Pruning}}      & $b$ & 0.01 & 0.01 & 0.001 & 0.001 \\
    & $N_C$ & 64 & 64 & 64 & 64 \\
    & $D$ & 128 & 128 & 128 & 128 \\
    & PFB Location & stage-1 & stage-1 & stage-2 & stage-1 \\
    & epoch start pruning & 5 & 5 & 5 & 15 \\
    & epoch stop pruning & 180 & 180 & 80 & 265 \\
    \bottomrule
    \end{tabular}
    }
\caption{Detailed training settings on image classification datasets.}
\label{tab:appendix_details}
\end{table} 

\end{document}